\pdfoutput=1

\documentclass[11pt]{article}

\usepackage[]{emnlp2021}

\usepackage{times}
\usepackage{latexsym}

\usepackage[T1]{fontenc}

\usepackage[utf8]{inputenc}
\usepackage{times}
\usepackage{latexsym}
\usepackage{graphicx}
\usepackage{lipsum}

\usepackage{float}
\usepackage{algorithm}
\usepackage{algorithmic}
\usepackage{placeins}
\usepackage{subfigure}
\usepackage{booktabs}
\usepackage{cleveref}
\usepackage{multirow}
\usepackage{microtype}

\title{How Does Fine-tuning Affect the Geometry of Embedding Space:

A Case Study on Isotropy}

\author{Sara Rajaee$^\diamondsuit$ \and Mohammad Taher Pilehvar$^\spadesuit$ \\
  $^\diamondsuit$ Iran University of Science and Technology, Tehran, Iran \\
  $^\spadesuit$ Tehran Institute for Advanced Studies, Khatam University, Tehran, Iran \\
  \texttt{sara{\_}rajaee@comp.iust.ac.ir} \\
  \texttt{mp792@cam.ac.uk}}

\begin{document}
\maketitle
\begin{abstract}
It is widely accepted that fine-tuning pre-trained language models usually brings about performance improvements in downstream tasks.
However, there are limited studies on the reasons behind this effectiveness, particularly from the viewpoint of structural changes in the embedding space.
Trying to fill this gap, in this paper, we analyze the extent to which the isotropy of the embedding space changes after fine-tuning. 
We demonstrate that, even though isotropy is a desirable geometrical property, fine-tuning does not necessarily result in isotropy enhancements.
Moreover, local structures in pre-trained contextual word representations (CWRs), such as those encoding token types or frequency, undergo a massive change during fine-tuning.
Our experiments show dramatic growth in the number of elongated directions in the embedding space, which, in contrast to pre-trained CWRs, carry the essential linguistic knowledge in the fine-tuned embedding space, making existing isotropy enhancement methods ineffective.    
\end{abstract}

\section{Introduction}
Recently, several studies have focused on the remarkable potential of pre-trained language models, such as BERT \citep{devlin-etal-2019-bert}, in capturing linguistic knowledge. 
They have shown that pre-trained representations are able to encode various linguistic properties \citep{tenney-etal-2019-bert,talmor-etal-2020-olmpics,goodwin-etal-2020-probing,wu-etal-2020-perturbed,zhou-srikumar-2021-directprobe,DBLP:conf/iclr/ChenFXXTCJ21,DBLP:conf/iclr/TenneyXCWPMKDBD19}, among others, syntactic, such as part of speech \citep{liu-etal-2019-linguistic} and dependency tree \citep{hewitt-manning-2019-structural}, and semantic, such as word senses \citep{NEURIPS2019_159c1ffe} and semantic dependency
\citep{10.1162/tacl_a_00363}.

Despite their significant potential, pre-trained representations suffer from important weaknesses. 
Frequency and gender bias are two well-known problems in CWRs. While the former hurts the semantic expressiveness of embedding space, the latter reflects the unwanted social bias in training data ~\citep{li-etal-2020-sentence,DBLP:journals/pnas/GargSJZ18,gonen-goldberg-2019-lipstick}.
 The representation degeneration problem is another issue that limits their linguistic capacity.
~\citet{gao2018representation} showed that the weight tying trick ~\citep{DBLP:conf/iclr/InanKS17} in the pre-training procedure is mainly responsible for the degeneration problem in the embedding space. 
In such a case, the embeddings occupy a narrow cone in the space ~\citep{ethayarajh-2019-contextual}.
Several approaches have been proposed to improve the isotropy of pre-trained models, which in turn boosts the representation power and downstream performance of CWRs ~\citep{zhang-etal-2020-revisiting,Wang2020Improving}.
However, previous studies have mainly focused on the anisotropy of pre-trained language models.
Here, we investigate the impact of fine-tuning on isotropy. Specifically, We try to answer the following questions:
\begin{itemize}
    \item Can the improved performance achieved by fine-tuning pre-trained language models (LMs) be attributed to the increased isotropy of the embedding space?
    \item Does isotropy enhancement (using methods that null out dominant directions) have the same positive outcome for the fine-tuned models as it has for the pre-trained ones?
    \item How does the distribution of CWRs change upon fine-tuning?
\end{itemize}

To answer these questions, we consider the semantic textual similarity (STS) as the target task and leverage the metric proposed by \citet{mu2018allbutthetop} for measuring isotropy.
The pre-trained BERT and RoBERTa \citep{Liu2019RoBERTaAR} underperform static embeddings on STS, while fine-tuning significantly boosts their performance, suggesting the considerable change that CWRs undergo during fine-tuning ~\citep{reimers-gurevych-2019-sentence,rajaee-pilehvar-2021-cluster}.

Our analysis on the fine-tuned embedding space of BERT and RoBERTa demonstrates that word representations are highly anisotropic across all layers. An evaluation specifically carried out on the [CLS] tokens approves a similar pattern but to a greater extent.
Moreover, experimental results show fading of local clustered areas in pre-trained CWRs during fine-tuning, which could be a possible reason for the improved performance. 
Interestingly, the fine-tuning procedure can change the linguistic knowledge encoded in dominant directions of embedding space from unnecessary information to the essential knowledge for the target task such that eliminating them toward making isotropic space hurts the performance of contextual representations.

\section{Related Work}

Following the research line in understanding the reasons behind the outstanding performance of pre-trained language models and their capabilities, most recent investigations on fine-tuning have been done through probing tasks and by evaluating the encoded linguistic knowledge ~\cite{merchant-etal-2020-happens,mosbach-etal-2020-interplay,talmor-etal-2020-olmpics,yu-ettinger-2021-interplay}. 
These studies demonstrate that most changes in fine-tuning are applied to the upper layers, such that those layers encode task-specific knowledge, while lower layers are responsible for the core linguistic phenomenon \cite{durrani-etal-2021-transfer}. Moreover, the results show that some linguistic information is surprisingly eliminated by this procedure ~\citep{mosbach-etal-2020-interplay}. Studies on the multi-head attention structure suggest a similar trend in their patterns during fine-tuning; in higher layers, attention weights experience more significant changes \cite{hao-etal-2020-investigating}. More detailed analysis on self-attention modules indicates that dense and value projection matrices have heavily been affected by fine-tuning \cite{pmlr-v108-radiya-dixit20a}.
However, geometric analysis on the embedding space and changes applied to the structure of embeddings during fine-tuning are aspects that have not been properly understood. Furthermore, evaluating the fine-tuning effect on frequency bias in CWRs is another aspect that distinguishes our work from previous studies.

\section{Methodology}

\subsection{Background}
\paragraph{Fine-tuning} is a straightforward yet quite effective process for taking advantage of the linguistic knowledge encoded in pre-trained models and for achieving high performance on different downstream tasks ~\citep{peters-etal-2019-tune}. 
The [CLS] embedding or other strategies in calculating sentence representations (e.g., max- or mean-pooling) can be considered as the input to the classifier layer, which is jointly trained with the parameters of the pre-trained model on a specific task ~\citep{devlin-etal-2019-bert}.

\paragraph{Isotropy} is a geometrical assessment of the distribution of data points in a feature space, which is ideally uniform ~\citep{gao2018representation}.
An embedding space is considered isotropic if the word embeddings are not biased towards a specific direction (feature). In other words, in isotropic space, word embeddings are uniformly distributed in the space, leading to low correlation and near-zero cosine similarity for randomly sampled words. 

Contextual embedding spaces are known to lack the desirable isotropy property \citep{rajaee-pilehvar-2021-cluster,ethayarajh-2019-contextual}. 
~\citet{gao2018representation} called the defect \emph{the representation degeneration problem} and attributed it mainly to the weight tying trick \citep{press-wolf-2017-using} and the language modeling as the objective of the training. 
Under such a circumstance, random word embeddings are highly similar to one another while shaping a narrow cone in the space. Clearly, anisotropic distribution hurts the expressiveness of the embedding space, especially for semantic downstream tasks.

Cosine similarity-based metrics have usually been employed for assessing the isotropy of embedding spaces where a near-zero cosine similarity between random embeddings indicates isotropic distribution. 
However, \citet{rajaee-pilehvar-2021-cluster} demonstrated that these metrics might not be reliable for calculating isotropy since, in some cases, the cosine similarity of random words is zero while their distribution is not uniform. 
Hence, we utilize another metric based on Principal Components (PCs). 

As we mentioned before, anisotropic embedding spaces have unusual elongations toward different directions. 
Using the eigenvectors calculated in Principal Component Analysis (PCA) procedure, we can find the most elongated directions of the space, which are the reason for anisotropic distribution. 
The distribution is more uniform and isotropic if the extent of elongation is similar across different directions (the most and the least elongated directions). 
With this in mind, ~\citet{mu2018allbutthetop} proposed a measurement to quantify the embedding space isotropy employing PCs as follows:
\begin{equation}
    I(\mathcal{W}) = \frac{min_{u\in U}F(u)}{max_{u\in U}F(u)}\label{eq:1}
\end{equation} 

\noindent where $U$ is the set of all eigenvectors of the word embedding matrix, and $F(u)$ is the following partition function: 

\begin{equation}
    F(u) = \sum_{i = 1}^{N} e^{u^T w_i}\label{eq:2}
\end{equation}

\noindent where $N$ is the number of word embeddings and $w_i$ is the i$^{th}$ word embedding.
~\citet{arora-etal-2016-latent} demonstrated that for a perfectly isotropic embedding space, $F(u)$ could be approximated by a constant. 
The value of $I(\mathcal{W})$ is closer to one for the more isotropic embedding spaces.

\subsection{Methodology}
\label{methods}
We study the changes applied to the embedding space by fine-tuning from the perspective of isotropy. In this regard, we take several approaches explained as follows. 

\paragraph{Zero-mean.} This method simply transfers all the embeddings to the center. 

\paragraph{Clustering+ZM.} Here, we first cluster embeddings and then separately make each cluster zero-mean \citep{cai2021isotropy}.

These two approaches give us a precise picture of the extent of isotropy in the fine-tuned embedding space, globally and locally, since making zero-mean is a prerequisite for measuring isotropy \citep{mu2018allbutthetop}.

\paragraph{Global app.} This is a simple and effective post-processing algorithm for improving the isotropy of embedding space proposed by \citet{mu2018allbutthetop}. In this method, after making embeddings zero-mean, a few top dominant directions calculated using PCA are being discarded. 

\paragraph{Cluster-based app.} Based on the clustered structure of pre-trained LMs \citep{michael-etal-2020-asking,NEURIPS2019_159c1ffe}, this method can significantly improve the performance of contextual embedding spaces as well as their isotropy \citep{rajaee-pilehvar-2021-cluster}. Here, we first cluster embeddings and then make each cluster zero-mean individually. At the last step, dominant directions are calculated in each cluster and discarded. 

The last two approaches help us make the embedding space isotropic and potentially attain performance improvement. 
Moreover, they give us an insight into the changes of clustered structure of pre-trained models during fine-tuning.

\begin{table*}[ht]
\centering
\scalebox{0.90}{

\begin{tabular}{l c@{\hspace{3pt}}c c@{\hspace{3pt}}c c@{\hspace{3pt}}c c@{\hspace{3pt}}c c@{\hspace{3pt}}c c@{\hspace{3pt}}c}

\toprule
\multicolumn{1}{c}{}                         & 
\multicolumn{2}{c}{\textbf{Baseline}}        &
\multicolumn{2}{c}{\textbf{Zero-mean}}       &
\multicolumn{2}{c}{\textbf{Clustering+ZM}}   &
\multicolumn{2}{c}{\textbf{Global}} &
\multicolumn{2}{c}{\textbf{Cluster-based}}  \\
\cmidrule(lr){2-3}
\cmidrule(lr){4-5}
\cmidrule(lr){6-7}
\cmidrule(lr){8-9}
\cmidrule(lr){10-11}
& Perf. & Isotropy & 
Perf. & Isotropy & 
Perf. & Isotropy & 
Perf. & Isotropy & 
Perf. & Isotropy \\
\midrule

\textbf{Pre-trained}$^\dag$     & ~54.14 & 1.1E-5~     & ~59.70 & 1.1E-4~    & ~~67.73 
                         & 0.31   
                         & ~69.20 & 0.59~       & ~\textbf{74.01}     & \textbf{0.83}~   \\ 
\textbf{Fine-tuned}$^\dag$       & ~84.41 & 4.1E-3~     & ~\textbf{84.94}     & 6.6E-3~  
                         & ~~80.10 & 0.11       & ~82.14 & 0.22~      & ~64.43 
                         & \textbf{0.60}~ \\
\midrule
\textbf{Pre-trained}$^\ddag$      & ~33.99 & 2.5E-6~     & ~37.66 & 8.3E-2~    & ~~60.32 
                         & 0.69   
                         & ~65.99 & 0.86~       & ~\textbf{73.86}     & \textbf{0.95}~   \\ 
\textbf{Fine-tuned}$^\ddag$       & ~81.08 & 3.3E-4~     & ~\textbf{81.34}     & 6.1E-3~  
                         & ~~76.03 & 0.05       & ~79.71 & 0.18~      & ~60.96 
                         & \textbf{0.28}~ \\
\bottomrule
\end{tabular}
}

\caption{\label{experiment-tabel} 
Spearman correlation performance and isotropy for five different settings in the pre-trained and fine-tuned {BERT}$^\dag$ and {RoBERTa}$^\ddag$. Unlike the pre-trained models, increased isotropy does not bring about improved performance for the fine-tuned models.}
\end{table*}
\par

\subsection{Target Task}
To analyze changes in a fine-tuned model, we choose Semantic Textual Similarity (STS) as the target task considering STS-Benchmark dataset ~\citep{cer-etal-2017-semeval}. STS is a semantic regression task in which the model needs to determine the similarity of two sentences in a paired sample. The label is a continuous range in 0 to 5. 

The interesting point about STS, which makes it a reasonable choice for our analyses, is that the performance of pre-trained LMs is drastically low on this task ~\citep{reimers-gurevych-2019-sentence}. In fact, BERT and RoBERTa's contextual representations under-perform static embeddings, such as Glove ~\citep{pennington-etal-2014-glove} in this task.
Moreover, the [CLS] token, which is usually considered a sentence representation for classification tasks, has a lower performance than simple averaging over all tokens of a sentence. However, fine-tuning, whether with [CLS] token or mean-pooling method, can dramatically enhance the performance ~\citep{reimers-gurevych-2019-sentence}.

\subsection{Experimental Setup}
\label{sec-setups}
We analyze the influence of fine-tuning on the embedding space of the base versions of BERT and RoBERTa. 
Both models have similar transformer-based architectures, while RoBERTa has been trained with more training data and a slight difference in the optimization procedure. 
For the pre-trained setting, we use the models as feature extractors (the weights are frozen in this phase). Applying the mean-pooling method over the word embeddings, we obtain a sentence representation for every sample and consider the cosine similarity of the sentence representations as the textual similarity score. 
In the fine-tuning scenario, we fine-tune the models with a Siamese architecture introduced by ~\citet{reimers-gurevych-2019-sentence} that uses cosine similarity and the mean-pooling method for sentence representation.
In our experiments, the batch size is set to 32, the learning rate is set to 7E-5, and the models are fine-tuned for 3 epochs.
Following our previous work \citep{rajaee-pilehvar-2021-cluster}, we set the number of clusters and discarded dominant directions in Global and Cluster-based approaches to 27 and 12,  respectively, for both models.

\begin{table*}[ht]
\centering
\setlength{\tabcolsep}{11pt}
\scalebox{0.85}{

\begin{tabular}{l c@{\hspace{3pt}}c c@{\hspace{3pt}}c c@{\hspace{3pt}}c c@{\hspace{3pt}}c c@{\hspace{3pt}}c c@{\hspace{3pt}}c}
\toprule
\multicolumn{1}{c}{}                              & 
\multicolumn{2}{c}{}                              &
\multicolumn{4}{c}{\textbf{Global App.}}          &
\multicolumn{4}{c}{\textbf{Cluster-based App.}}   &\\
\cmidrule(lr){4-7}
\cmidrule(lr){8-11}

\multicolumn{1}{c}{}                              & 
\multicolumn{2}{c}{\textbf{Baseline}}             &
\multicolumn{2}{c}{\textbf{100 least dir.}}       &
\multicolumn{2}{c}{\textbf{700 least dir.}}       &
\multicolumn{2}{c}{\textbf{100 least dir.}}       &
\multicolumn{2}{c}{\textbf{700 least dir.}}  \\
\cmidrule(lr){2-3}
\cmidrule(lr){4-5}
\cmidrule(lr){6-7}
\cmidrule(lr){8-9}
\cmidrule(lr){10-11}
& Perf. & Isotropy & 
Perf. & Isotropy & 
Perf. & Isotropy & 
Perf. & Isotropy & 
Perf. & Isotropy \\
\midrule
\textbf{BERT}            & ~84.41  & 4.1E-3~     & ~84.93     & 2.2E-3~    & ~~82.93 
                         & 2.2E-3   
                         & ~77.87  & 0.10~       & ~75.10     & 0.16   \\ 
\textbf{RoBERTa}         & ~81.08  & 3.3E-4~     & ~{81.66}   & 3.2E-4~  
                         & ~~78.59 & ~1.4E-2~        & ~73.19     & 0.13~      & ~71.39 
                         & 0.13 \\

\bottomrule
\end{tabular}
}

\caption{\label{ling-tabel} 
Spearman correlation performance and isotropy after removing the least dominant directions in Global and Cluster-based approaches on STS dev set. The results suggest the low sensitivity of the fine-tuned models to eliminating more than 90\% of directions with lower elongations. }
\end{table*}

\section{Findings}
\begin{figure}[t!]
    \centering
    
    \includegraphics[width=8.2cm,height=7.2cm]{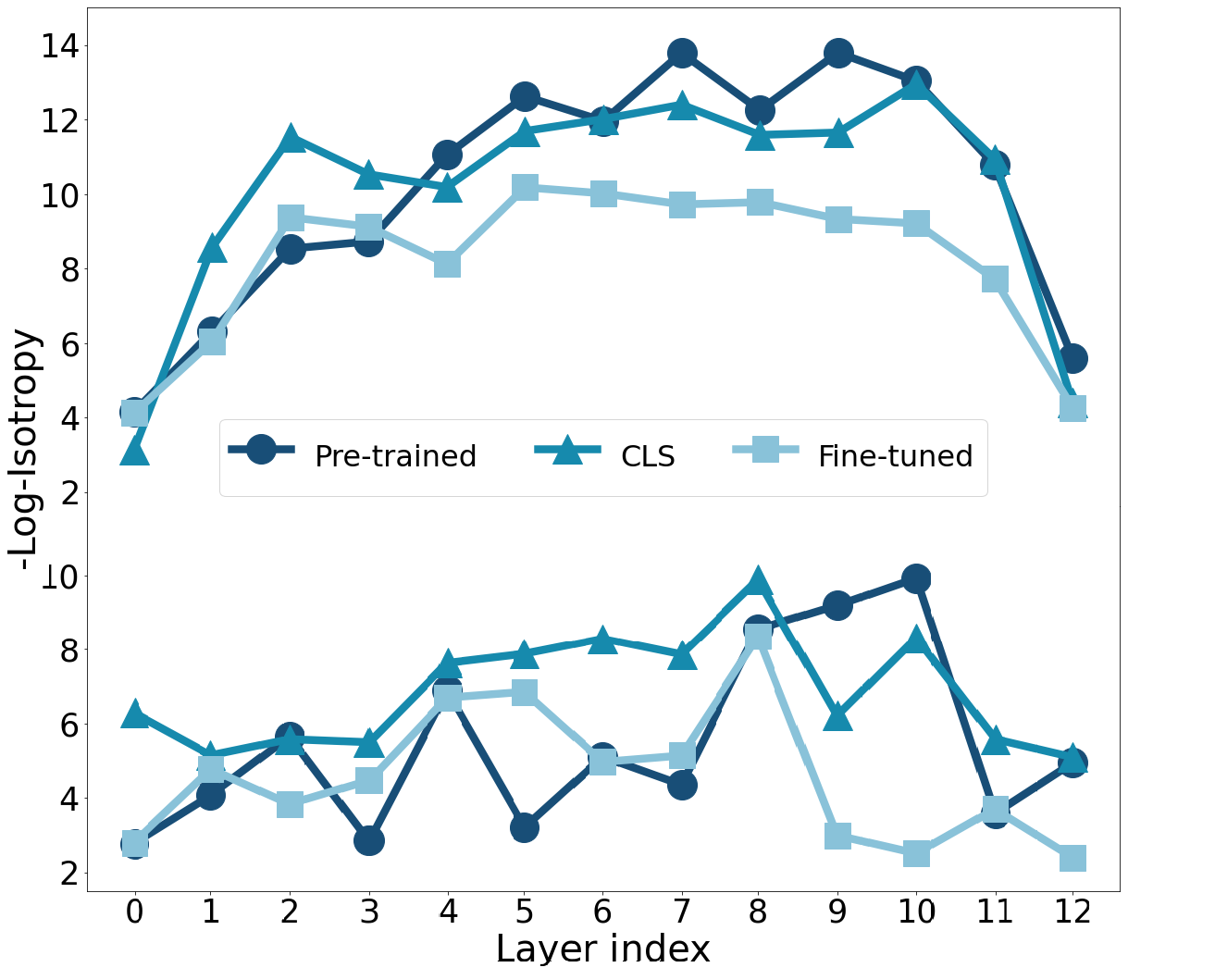}
    
    \caption{Negative \textit{log} of isotropy for [CLS] tokens, and all the tokens in the pre-trained and fine-tuned embedding space in all layers of BERT (bottom) and RoBERTa (top) using $I(\mathcal{W})$ on STS-B dev set. Higher values indicate lower isotropy.}
    \label{fig:fig-iso-alllayers}
\end{figure}

\paragraph{The embedding space of fine-tuned models is still highly anisotropic.}
Figure \ref{fig:fig-iso-alllayers} depicts our experimental results on evaluating the isotropy in the models’ embedding spaces using $I(\mathcal{W})$. We take the pre-trained embedding space as a baseline and compare its isotropy to the fine-tuned space (all representations) and the [CLS] tokens in all layers.
The results demonstrate that performance enhancements achieved after fine-tuning cannot be attributed to the increased isotropy of the embedding space. 
Although fine-tuning improves isotropy, specifically in the upper layers, the distribution of embeddings is still highly non-uniform. 
Moreover, in most layers, the [CLS] tokens' representations are much more anisotropic than all representations in the fine-tuned space. These patterns hold for both BERT and RoBERTa, while the latter tends to be more anisotropic. 
We also note that although different random seeds change the reported numbers, the difference between isotropy of [CLS], fine-tuned, and pre-trained embedding spaces remain.

\paragraph{Adjusting the fine-tuned embedding space for isotropy hurts its performance.}
Several studies have shown that isotropy has theoretical and practical benefits. A natural question that arises here is if increasing the isotropy of a fine-tuned embedding space would lead to further performance improvements?
To examine this hypothesis, we fine-tuned the models with the Siamese architecture and considered the settings explained in Section \ref{methods}.
Results are listed in Table \ref{experiment-tabel}. 
Clearly, as opposed to pre-trained models, increasing isotropy of the fine-tuned embedding space does not enhance performance. 
Instead, we observe a drop in performance.
This can be attributed to the fact that fine-tuning concentrates information about the target task in the dominant directions, whether it is obtained during the fine-tuning procedure or just brought up from the encoded knowledge in the pre-trained model. 

\begin{figure}[t!]
    \centering
    \includegraphics[width=8cm,height=6.9cm]{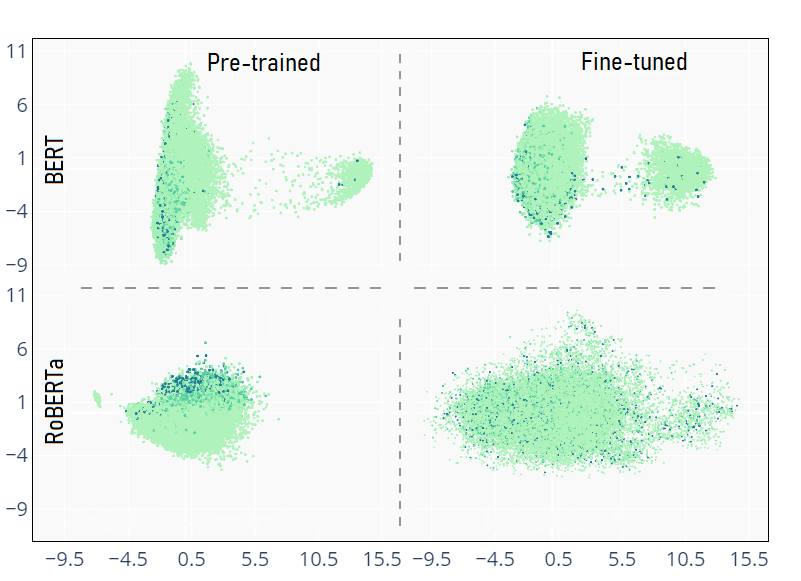} 
    \caption{Illustration of pre-trained and fine-tuned CWRs colored based on their frequency in BERT and RoBERTa (using Wikipedia dump as corpus). The more frequent words have darker colors. As can be observed, the embedding space is still anisotropic after fine-tuning, while the frequency-based distribution of CWRs has been remedied.}
    \label{fig:fig-word-freq}

\end{figure}

\paragraph{The fine-tuned models heavily rely on a few top directions to solve the target task.}
To investigate the sensitivity of the fine-tuned model to the linguistic knowledge encoded in different directions, we discarded the least dominant directions and evaluated the performance of representations. 
The results of the experiment have been presented in Table \ref{ling-tabel}.
By eliminating the 100 and 700 least dominant directions from a total of 768 directions, we observe a slight drop in the performance compared to removing 12 top dominant directions.
This suggests that the top dominant directions carry essential knowledge about the target task. 
We leave further investigation of this interesting behavior to future work.

\paragraph{The clustered structure of the embedding space changes during fine-tuning.} 
The results of the Clustering+ZM setting and Cluster-based approach in Table \ref{experiment-tabel} show that the clustered structure of the pre-trained embedding space \citep{cai2021isotropy} has faded in the fine-tuned CWRs. These two settings can improve the STS performance of the pre-trained model by increasing isotropy. However, applying them to fine-tuned CWRs leads to performance reduction.
Moreover, as can be seen in Figure \ref{fig:fig-word-freq}, the local areas that encode frequency information in the pre-trained CWRs have been removed by fine-tuning, which can be a reason for the high performance of fine-tuned representations.

\paragraph{The number of elongated dominant directions significantly increases after fine-tuning.} The results of Global and Cluster-based approaches in Table \ref{experiment-tabel} reveal that with equal numbers of removed directions, the fine-tuned embedding space is less isotropic compared to the pre-trained one. This means that to have similar embedding spaces in terms of isotropy, we need to eliminate more dominant directions from the fine-tuned embedding space.

\section{Conclusions}

In this paper, we explored the effect of fine-tuning on the structure of the embedding space of BERT and RoBERTa. 
Our analysis demonstrates that the remarkable performance usually gained as a result of fine-tuning is not due to its enhancement of isotropy in the embedding space.
Similarly to their pre-trained counterparts, fine-tuned CWRs have elongated directions towards different dimensions across all layers, and the number of these directions tends to increase by fine-tuning. 
We have also found that fine-tuning changes the nature of the linguistic knowledge encoded in dominant directions such that removing them hurts the performance (unlike pre-trained models for which removing such directions often result in performance improvements). 
Moreover, the clustered structure of pre-trained models is entirely modified upon fine-tuning, producing unbiased embedding space from the viewpoint of word frequency. 

As future work, we plan to experiment with more target tasks and different fine-tuning strategies to expand our knowledge about the fine-tuning procedure.
Furthermore, we aim at exploring the type of linguistic knowledge encoded in specific dimensions or subspaces in the semantic space.


\bibliography{anthology,emnlp2021}
\bibliographystyle{acl_natbib}

\end{document}